\documentclass[10pt,twocolumn,letterpaper]{article}

\usepackage{iccv_modifiedforworkshop}
\usepackage{times}
\usepackage{epsfig}
\usepackage{graphicx}
\usepackage{amsmath}
\usepackage{amssymb}

\usepackage[pagebackref=true,breaklinks=true,letterpaper=true,colorlinks,bookmarks=false]{hyperref}

\usepackage{cleveref}

\iccvfinalcopy 

\def\iccvPaperID{****} 

\ificcvfinal\fi

\begin{document}

\title{Language as the Medium: Multimodal Video Classification through text only}

\author{Laura Hanu\\
\and
Anita L. Verő\\
Unitary\\
{\tt\small \{laura,anita,james\}@unitary.ai}
\and
James Thewlis\\
}
\maketitle
\ificcvfinal\thispagestyle{empty}\fi

\begin{abstract}
    Despite an exciting new wave of multimodal machine learning models, current approaches still struggle to interpret the complex contextual relationships between the different modalities present in videos. Going beyond existing methods that emphasize simple activities or objects, we propose a new model-agnostic approach for generating detailed textual descriptions that captures multimodal video information. Our method leverages the extensive knowledge learnt by large language models, such as GPT-3.5 or Llama2, to reason about textual descriptions of the visual and aural modalities, obtained from BLIP-2, Whisper and ImageBind. Without needing additional finetuning of video-text models or datasets, we demonstrate that available LLMs have the ability to use these multimodal textual descriptions as proxies for ``sight'' or ``hearing'' and perform zero-shot multimodal classification of videos in-context. Our evaluations on popular action recognition benchmarks, such as UCF-101 or Kinetics, show these context-rich descriptions can be successfully used in video understanding tasks. This method points towards a promising new research direction in multimodal classification, demonstrating how an interplay between textual, visual and auditory machine learning models can enable more holistic video understanding.
\end{abstract}

\begin{figure*}
\centering
\includegraphics[width=0.82\linewidth]{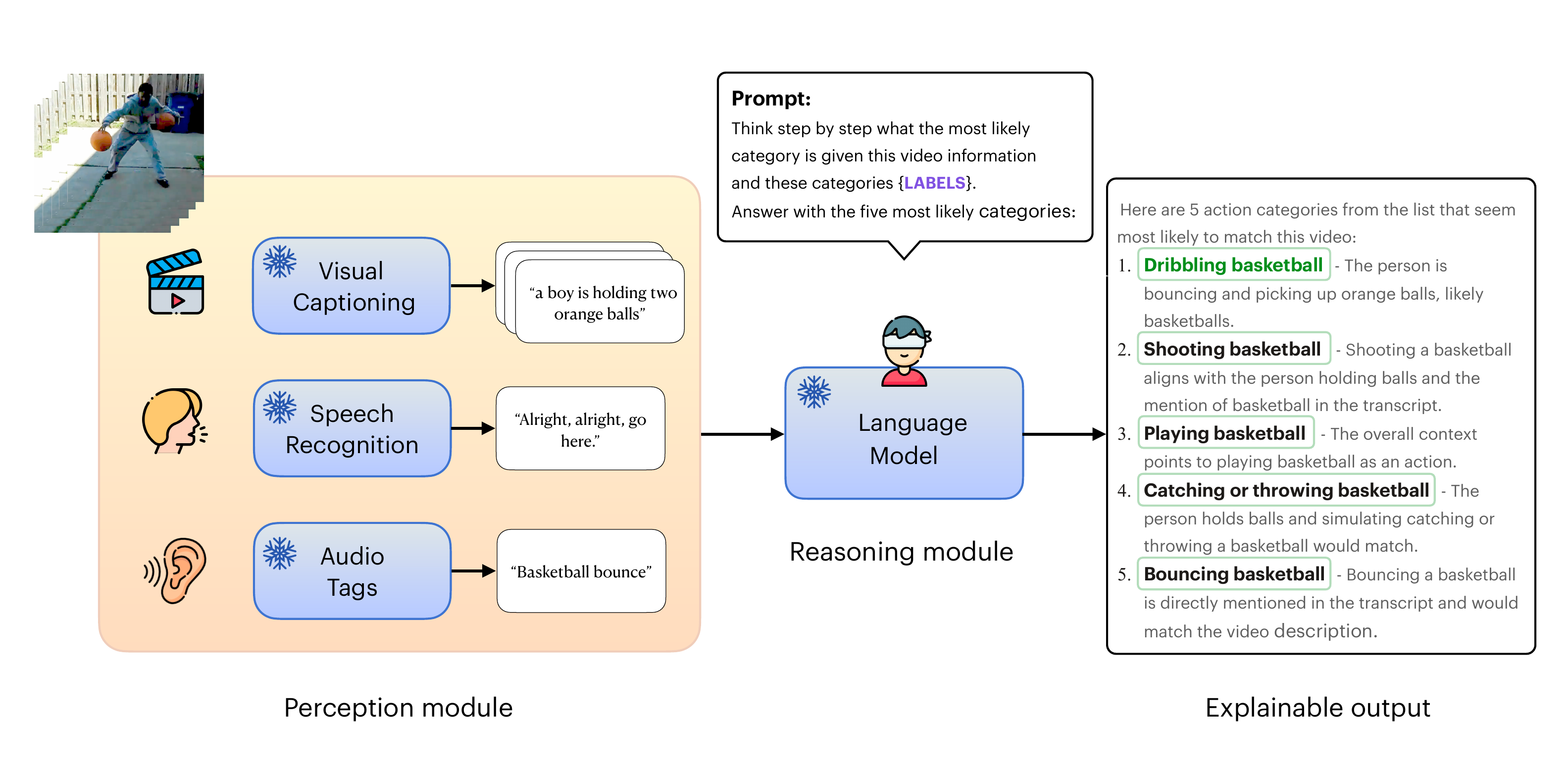}
\caption{Our method combines a ``perception'' module, which uses visual and auditory models to get multimodal textual descriptors as sensory proxies for ``sight'' and ``hearing'', and a ``reasoning'' module that processes these textual inputs to form a coherent narrative and identify the likeliest content in the video, completed by justifications.}
\label{fig:label}
\end{figure*}

\section{Introduction}

Imagine it is the year 2008 and you have just watched the latest episode of Breaking Bad - a highly multimodal experience
featuring moving pictures, speech and sound effects. Suddenly you receive a text message on your mobile phone -- it is your colleague, who is urgently requesting a description of the episode so that they may participate in water cooler discussions without arousing suspicion.

You must now convey to your colleague, using only text messages, a description of the episode that will stand up to scrutiny. Although reducing the vast amount of pixels and audio samples you have just consumed down to a few words seems like a daunting task, you recognize that by combining succinct descriptions of key images, speech and sounds with your colleague's inherent ability to fill in gaps using contextual reasoning you will be able to provide a comprehensive recount of the episode without the need for the direct experience. In this work, we explore to what extent Large Language Models (LLMs) are able to perform
a similar task, namely classifying the action in videos when receiving only textual clues about the video contents from other models.

The last few years have seen remarkable progress in large language models for text, which have shown unprecedented capabilities and performance on downstream tasks \cite{brown2020language,scao2022bloom,openai2023gpt4,touvron2023llama}. This has led to methods trying to bridge the gap between vision and language. Contrastive methods such as CLIP train joint vision language representations \cite{radford2021learning}. Perceiver IO \cite{jaegle2021perceiver} offers a generic scheme to encode arbitrary modalities. Kosmos \cite{huang2023language} is a large multimodal model trained from scratch on web-scale image and text data. GPT-4 \cite{openai2023gpt4} accepts image input, but this feature is not currently publicly available. Numerous works adapt pretrained LLMs in order to understand information from different modalities. Flamingo \cite{alayrac2022flamingo} injects representations of images
and short videos into the language model backbone
using Gated X-attn layers. BLIP-2 \cite{li2023blip} introduces a Q-Former which provides 
``soft visual prompts'' to condition an LLM on visual information. Mini-GPT4 \cite{zhu2023minigpt} leverages the Q-Former to provide a soft prompt to a Llama-based model. In contrast to these techniques, we demonstrate that using only text as the medium can convey multimodal information to downstream LLMs. This has several key advantages. Firstly, this approach ensures a straightforward ``plug and play'' interface for chaining models without extra adaptation. This is particularly relevant with the rise of API-based language models that prohibit modifications. Secondly, inter-model communication becomes transparent and interpretable in natural language. Crucially, this method simplifies tasks like multimodal video classification into two phases: a ``perception'' phase using unimodal or multimodal models as surrogates for various senses, followed by a ``reasoning'' phase where a foundation model consolidates diverse inputs to create a comprehensive video narrative.

More recent methods such as LENS \cite{berrios2023towards} or Video ChatCaptioner \cite{chen2023video} explore similar textual interactions between models. While LENS only explores the ability of LLMs to reason over visual question answering tasks given visual clues about images, Video ChatCaptioner proposes chaining together BLIP-2 and ChatGPT in order to have conversations about images. Our method goes beyond just question answering tasks, demonstrating that both visual and auditory clues can be used by LLMs for video classification.

In summary, our contributions are: 1) We introduce a new multimodal classification approach consisting of two phases: a ``perception" phase where models act as sensory proxies and a ``reasoning" phase that consolidates multimodal textual inputs into a coherent narrative. 2) We demonstrate the efficacy of text as the primary medium of interpreting multimodal data. 3) For the first time, we showcase that textual representations of visual and auditory cues alone can effectively classify actions within videos.

\section{Method}

\paragraph{Perception models}
To extract visual captions from video frames, we use the BLIP-2 \cite{li2023blip} model. We process only 5 equidistant frames per video to ensure a diverse sampling of the video content. We use Whisper \cite{radford2023robust} to obtain audio transcripts, specifically the Faster Whisper version \cite{fastwhisper} which has been optimised for fast inference. We use a temperature of 0, a beam size of 5 and the VAD filter to exclude the parts of the video that don't have any speech. In order to generate audio tags, we leverage ImageBind \cite{girdhar2023imagebind} to get audio embeddings and compute the similarity with the textual embeddings of the AudioSet labels. We then only select the labels that have a similarity over a certain threshold, which can be obtained by qualitatively checking a few examples. 

\paragraph{Reasoning models}

For our reasoning module, we test out 3 different state-of-the-art large language models. We use the GPT completion API, specifically the GPT3.5-turbo version \cite{gpt}.  Additionally, we use the newly launched function calling feature which allows the user to specify a json schema for the output. The second LLM we evaluate is Claude-instant-1 \cite{claude}, which has reported similar performance and capabilities to GPT3.5-turbo. For Llama2 we use the \texttt{Llama-2-13b-chat} variant \cite{touvron2023llama}, which has 13 Billion parameters and is specialised for conversation. We use a temperature of 0 or near 0 for all reasoning models to ensure more consistent outputs that are able to better adhere to the instructions given. The prompts we use for classification follow this simple template with slight variations among the different LLMs to accomodate specific prompt guidelines: \textit{Given this \{multimodal clues\} and these action recognition labels: \{labels\} Please return the 5 labels that apply the most to the video in a json format, from the most likely to the least likely.}

\paragraph{Structured Output}

LLMs usually generate free-flowing natural language outputs, however, for the task of classification we want the
model to provide us with 5 ranked guesses from a set of pre-defined class names. To accomplish this with GPT we use the function calling API, providing the model with a JSON Schema of the function to call, where the schema contains an enum of the possible class names.

For Claude, we provide the class names in the prompt and ask for the results to be returned as JSON, which, in the majority of cases, results in a JSON object with a ``labels'' key containing a list of most likely labels, or an object whose keys are the classes and values are the rank.

For Llama2, we provide class names in the System prompt, and observe that predictions are usually included as a numbered list in the output, hence we simply parse lines beginning with a number. To compare with ground truth, we normalise to remove spaces and convert to lowercase.

For all models, occasionally the output cannot be parsed (such as hallucinated class names or extra characters), and in this case we consider the prediction to be incorrect.

\section{Evaluation Datasets}

\paragraph{UCF101} The UCF-101 test set comprises of 13,320 short video clips from YouTube spanning 101 action categories, providing a diverse set of everyday human actions, ranging from playing instruments to sports activities.

\paragraph{Kinetics400} The Kinetics400 test set contains 10s Youtube video clips and 400 human action classes. In order to circumvent API costs, since the test set contains 38,685 video clips, we construct a smaller representative subset of 2000 videos, sampling 5 videos per category.

\section{Experiments}

\begin{table}[t]

    \footnotesize
    \centering
    \setlength{\tabcolsep}{3pt}
    \begin{tabular}{lcccc}
     &  
    \multicolumn{2}{c}{\textbf{UCF-101}} & \multicolumn{2}{c}{\textbf{K400*}} \\
    \hline
    \textbf{Model} & \textbf{1-Acc.} &  \textbf{5-Acc.} &  \textbf{1-Acc.} & \textbf{5-Acc.} \\
    \hline
    BLIP2+Claude-1(caps) & 63.01  & 85.35 & 38.90 & 54.20\\
    BLIP2+Claude-1(caps, speech) & 67.06 & 86.13 & 41.20 & 57.00\\
    BLIP2+Claude-1(caps, speech, audio) & 67.13 & 86.15 & 41.20 & 57.35 \\
    \hline
    \end{tabular}
\caption{Comparing different levels of context on the UCF101 test set and a subset of Kinetics400 using Claude-instant-1.\label{tab:context}}
\end{table}

\begin{figure}[t]
\begin{center}
   \includegraphics[width=0.94\linewidth]{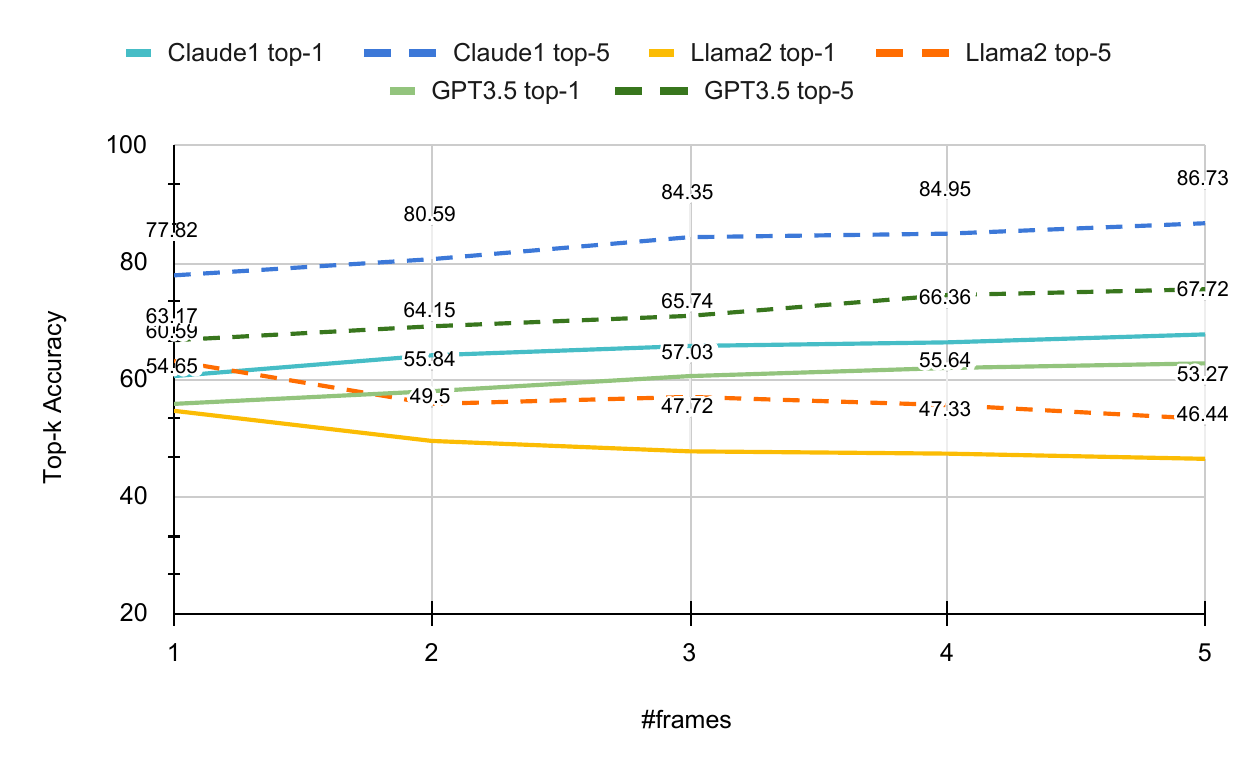}
\end{center}
   \caption{Comparing the ability of language models to filter context from captions corresponding to a varying number of frames. While Claude1 and GPT3.5 are able to leverage the extra information, it seems that the extra information overwhelms Llama2-13B.}
\label{fig:comparingframes}
\end{figure}

First, we run experiments to see the role of each modality in classifying videos on the UCF-101 test and the 2k subset of Kinetics400. As \Cref{tab:context} shows, the language model is able to benefit from additional audio information.
 In \Cref{tab:varyingllm} we compare how well the 3 different large language models used are able to interpret the visual and auditory information given. We find that both GPT3.5-turbo and Claude-instant-1 outperform Llama2, with Claude-instant-1 obtaining on average the highest accuracy.
 In Figure \ref{fig:comparingframes} we test the effect of including more or less frame captions. Interestingly, while both GPT3.5 and Claude-1 benefit from "seeing" more captions, LLama2's performance is negatively affected. We hypothesise this is due to the model becoming overwhelmed with the redundant information, making it more likely to pick a word from the captions rather than the label list.

\begin{figure*}
\centering
\includegraphics[width=0.9\linewidth]{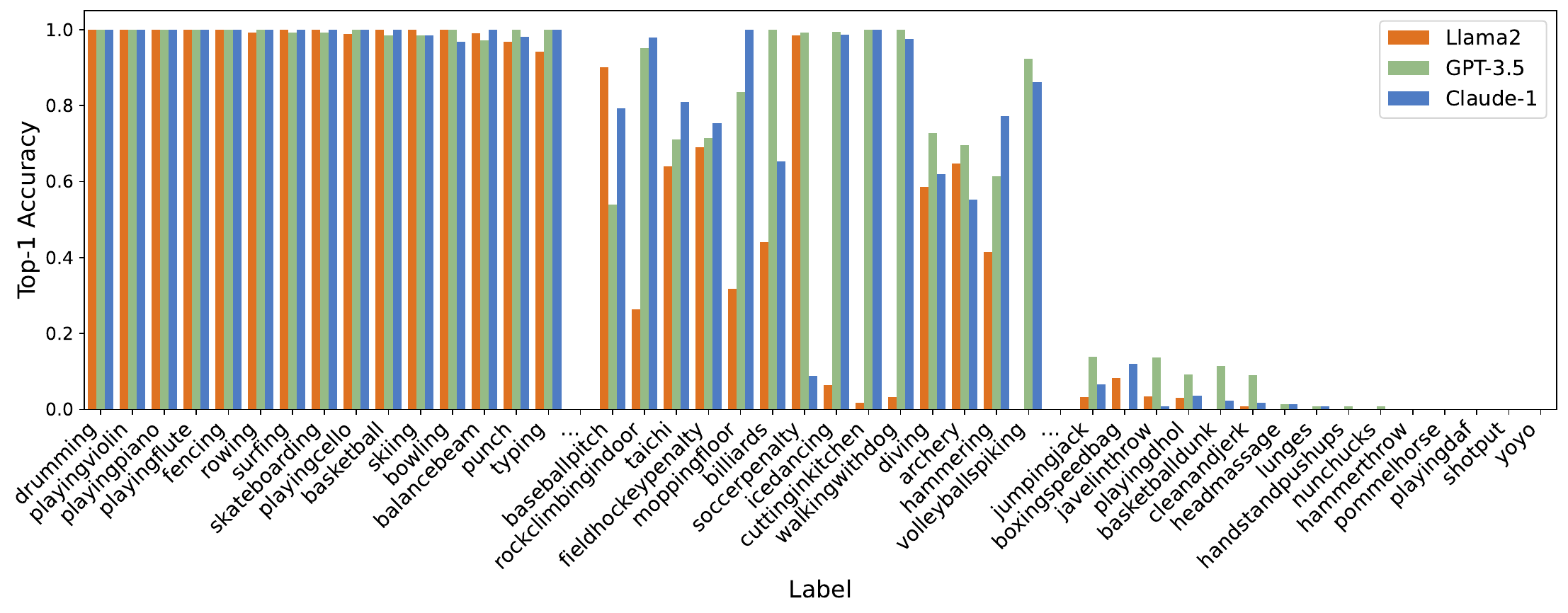}
\caption{Best and worst performing UCF-101 classes among the 3 language models used: Claude-instant-1, GPT3.5-turbo, Llama2-13B.}
\label{fig:accuracybylabel}
\end{figure*}

\begin{table}[t]
    \footnotesize
    \centering
    \setlength{\tabcolsep}{2pt}
    \begin{tabular}{|l|c|c|c|c|}
    \hline
    \textbf{Model} & \textbf{Top-1 acc.} & \textbf{Top-3 acc.} & \textbf{Top-5 acc.} \\
    \hline
    BLIP2(FlanT5-XXL)+Llama2-13B & 49.56 & 56.70 & 58.51 \\
    BLIP2(FlanT5-XXL)+GPT3.5 & 66.37 & 79.27 & 82.04 \\
    BLIP2(FlanT5-XXL)+Claude-1 & 63.01 & 81.49 &  85.35  \\
    \hline
    \end{tabular}
\caption{Comparing varying LLMs on the UCF101 test set. \label{tab:varyingllm}}
\end{table}

\section{Discussion and Future Work}

Using separate models to translate vision and speech into text might result in the inability to model interactions between modalities, resulting in an incomplete context. Image analysis on a frame by frame basis also lacks proper temporal modeling of persistent identities and relationships over time. Generative models are prone to hallucinations and unreliable outputs, therefore relying on them for complex tasks that require consistent outputs might prove challenging in practice. Moreover, by only providing the text of the class names to the LLM, we are relying on the class names
being sufficiently descriptive. For example, UCF-101 contains classes related to very specific musical instruments
such as \texttt{PlayingDhol} or \texttt{PlayingTabla}, which would require the LLM to know that these
are percussion instruments and would also require the captioning model to be able to give enough
granular detail to distinguish these instruments.
As Figure \ref{fig:accuracybylabel} shows, the models struggle with actions which are harder to determine from frame information alone, such as \texttt{headmassage} or more obscure objects, such as \texttt{nunchuks} or \texttt{yoyo}. Failure modes usually include the inability of the visual captioning model to recognise the level of specificity needed to differentiate between similar actions.
Finally, even though our method is not quite competitive with the state of the art zero-shot performance of 91.5 Top-1 Accuracy on UCF-101 and 76.8 on Kinetics400 \cite{akbari2023alternating}, it is more generalizable to video understanding scenarios that require complex contextual reasoning.

Future work includes leveraging additional context, such as video comments \cite{hanu2022vtc} or employing a chat-based approach where the ``reasoning" module can ask the ``perception" module for clarification to get more information.

\section{Conclusion}
In this work, we have introduced a new framework for multimodal video classification that leverages text as the primary medium for combining signals across modalities. We demonstrate for the first time that chaining together perception models for vision, speech and audio with large language models can enable zero-shot video classification using only textual representations of multimodal signals. Our work highlights the potential of using natural language as a flexible interface for integrating signals across modalities.


{\small
\bibliographystyle{ieee_initials}
\bibliography{egbib}
}

\end{document}